%% file: main.tex
\documentclass[runningheads]{llncs}

\usepackage{eccv}

\usepackage{eccvabbrv}

\usepackage{graphicx}
\usepackage{booktabs}
\usepackage{caption}
\usepackage{subcaption}

\usepackage{booktabs}
\usepackage{multirow}
\usepackage{tabularx}
\usepackage{array}
\usepackage{makecell}

\usepackage[accsupp]{axessibility}  % Improves PDF readability for those with disabilities.

\usepackage{hyperref}
\usepackage{bm}

\usepackage{orcidlink}

\begin{document}

% ---------------------------------------------------------------
% TODO REVIEW: Replace with your title
% \title{JacobianAvatar: Temporally Consistent Semi-rigid Avatar Generation from a Video}  % 2줄 
\title{JacobianAvatar: Temporally Consistent Semi-rigid Avatar Reconstruction from a Monocular Video}  % 3줄 

% TODO REVIEW: If the paper title is too long for the running head, you can set
% an abbreviated paper title here. If not, comment out.
\titlerunning{JacobianAvatar}

% TODO FINAL: Replace with your author list. 
% Include the authors' OCRID for the camera-ready version, if at all possible.
\author{Changyeon Won\inst{1}\orcidlink{0000-0001-5335-2606} \and
Min-Gyu Park\inst{2,3}\orcidlink{0000-0003-1752-150X} \and
Seonghwan Park\inst{2}\orcidlink{0009-0009-6322-6959}\and \\ 
Ju Hong Yoon\inst{2,3}\orcidlink{0000-0003-2945-8376}\and
Hae-Gon Jeon\*\inst{4}\thanks{Corresponding author}\orcidlink{0000-0003-1105-1666}
}

% TODO FINAL: Replace with an abbreviated list of authors.
\authorrunning{C. Won et al.}
% First names are abbreviated in the running head.
% If there are more than two authors, 'et al.' is used.

% TODO FINAL: Replace with your institution list.
% TODO FINAL: Replace with your institution list.
% \institute{
%     Department of AI Convergence, Gwangju Institute of Science and Technology \and 
%     Korea Electronics Technology Institute (KETI) \and  polygom \and
%     Department of Artificial Intelligence, Yonsei University \\ % 여기를 \and 대신 \\ 로 수정
%     \email{\tt\small cywon1997@gm.gist.ac.kr, 
%     \{mpark,shpark97,jhyoon\}@keti.re.kr, 
%     earboll@yonsei.ac.kr}
% }

\institute{
    \textsuperscript{1}GIST  \quad  \textsuperscript{2}KETI 
     \quad  \textsuperscript{3}polygom  
     \quad \textsuperscript{4} Yonsei University
    % \textsuperscript{1}Department of AI Convergence, Gwangju Institute of Science and Technology \\
    % \textsuperscript{2}Korea Electronics Technology Institute (KETI) \quad 
    % \textsuperscript{3}polygom \\
    % \textsuperscript{4}Department of Artificial Intelligence, Yonsei University \\
    % \email{\tt\small cywon1997@gm.gist.ac.kr,\\ \{mpark,shpark97,jhyoon\}@keti.re.kr,\\ earboll@yonsei.ac.kr}
}

\maketitle

\renewcommand\twocolumn[1][]{#1}%
\begin{center}
\includegraphics[width=0.9\linewidth]{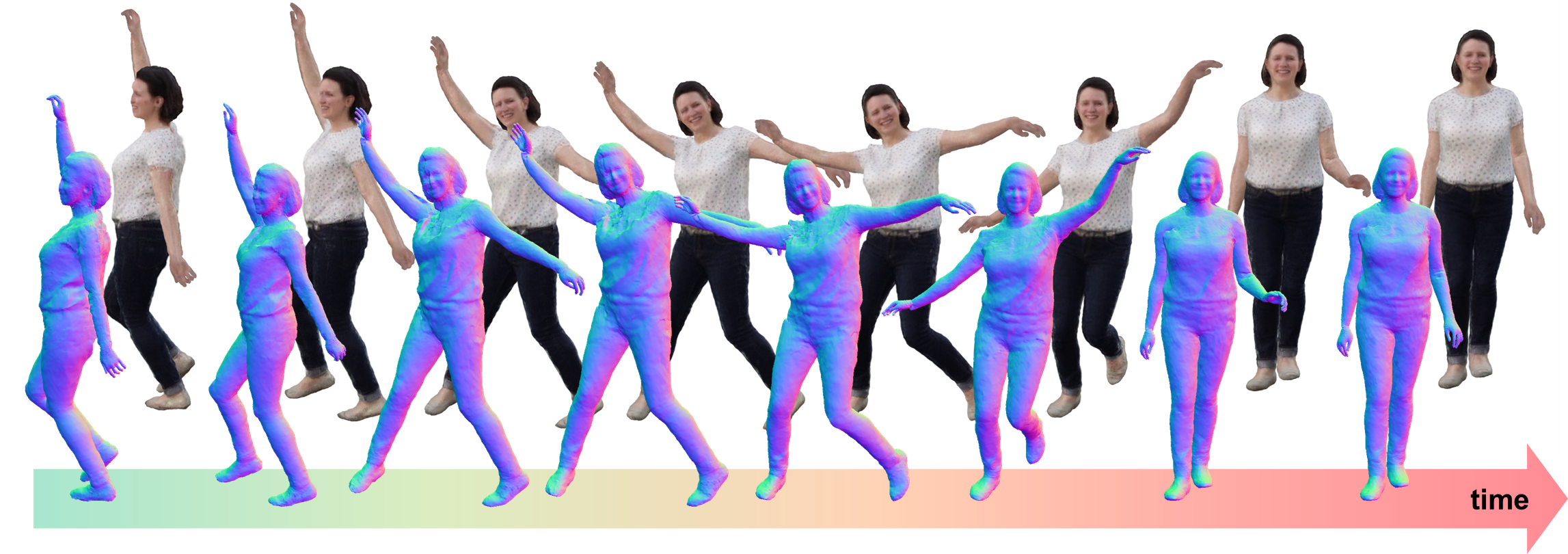}
    \captionof{figure}{JacobianAvatar. A neural representation for digital human avatars that captures rigid articulated motions and non-rigid local deformations using hierarchical neural Jacobian fields, while encouraging temporal consistency with high-fidelity geometry. The top and bottom rows show the rendered color images and normal maps of an animated avatar.
    }% In the upper figure, we visualize tracking paths of our JacobainAvatar. In bottom row illustrates the pose dependent semi-rigid deformations.
    %figure, we showcase the semi-rigid deformation which changes as pose changes. 
    \label{fig:teaser}
\end{center}

\input{sec_camera_ready/0_abstract}
\input{sec_camera_ready/1_intro}
\input{sec_camera_ready/2_related}
\input{sec_camera_ready/3_proposed}
\input{sec_camera_ready/4_experiment}

\input{sec_camera_ready/5_conclusion}
{
    \small
    \bibliographystyle{splncs04}
    \bibliography{main}
}
% WARNING: do not forget to delete the supplementary pages from your submission 
% \input{sec/X_suppl

 \clearpage
 \appendix % <--- 이것을 추가하면 이후 섹션은 A, B, C로 매겨집니다.
 \setcounter{table}{0} % 부록용 테이블 번호 초기화 (필요시)
 \setcounter{figure}{0} % 부록용 피규어 번호 초기화 (필요시)
 \renewcommand{\thetable}{\Alph{section}\arabic{table}} % A1, A2... 로 표시
 \renewcommand{\thefigure}{\Alph{section}\arabic{figure}}

 % Supplementary 제목을 예쁘게 넣고 싶다면 아래 코드를 추가하세요
 \begin{center}
     \Large \textbf{Supplementary Material}
\end{center}\input{sec_camera_ready/supp}

\end{document}

%% file: sec_camera_ready/0_abstract.tex
\begin{abstract}
Generating realistic human avatars in complex motions—such as clothing dynamics—requires modeling of global and local deformations which remains challenging in monocular settings. We address this problem by leveraging neural Jacobian fields (NJFs) for representing semi-rigid deformations. We train self-supervised neural networks for predicting Jacobian matrices that give the pose-dependent deformations, by solving a Poisson equation. However, monocular input presents several difficulties such as self-occluded regions and invisible surfaces. To address these issues, we introduce three key components: a constrained Poisson solver, signed distance-based Jacobian regularization, and a deformation-guided residual flow loss, which together suppress boundary artifacts, recover frequently occluded regions such as armpits and thighs, and enforce temporal consistency during motion. Experiments on benchmark and in-the-wild videos demonstrate that our method generates temporally stable and geometrically coherent avatars, outperforming state-of-the-art approaches.
\keywords{Avatar Reconstruction \and Temporal consistency \and Semi-rigid deformations}

\end{abstract}

%% file: sec_camera_ready/1_intro.tex
\section{Introduction}
\label{sec:intro}

Reconstructing photorealistic and animatable human avatars from a monocular video has received a huge attention, and recent advances have significantly improved the fidelity of clothed human modeling by adopting a semi-rigid deformation paradigm~\cite{codecavatar2025wang,D3human,exavatar}. This paradigm combines articulated skeletal motion governed by linear blend skinning (LBS)~\cite{SMPL:2015,SMPL-X:2019} with non-rigid local deformations to capture clothing wrinkles and pose-dependent surface details.

The semi-rigid reconstruction problem naturally decomposes into two tightly coupled subproblems. The first is the recovery of a canonical model that captures the subject’s intrinsic shape and appearance in a resting state. The second is the prediction of non-rigid local deformations, such as clothing wrinkles and pose-dependent correctives~\cite{codecavatar2025wang}, which is crucial for realistic rendering of an avatar in motion. The majority of the studies assume the body pose is the dominant source of causing local deformations~\cite{humannerf,jiang2022neuman,jiang2022selfrecon,vid2avatar,hu2024gaussianavatar,exavatar,song2023pose,lsavatar,facavatar}. This pose-driven approach, therefore, usually regresses local surface offsets conditioned on estimated human pose parameters~\cite{SMPL-X:2019}. Although these methods are effective in recovering pose-induced effects, they inherently overlook history-dependent deformations arising from factors beyond instantaneous pose, such as cloth inertia, contact, or motion velocity. Time-driven methods address this limitation by incorporating temporal context, modeling deformation as a function of joint velocities or sequential latent codes to better capture dynamic surface behaviors.
Since many approaches rely on Gaussian Splats~\cite{3dgs} and NeRF~\cite{humannerf} to represent a human avatar they heavily rely on the photometric reconstruction loss which does not encourage temporal consistency and geometric details, though rendered images look plausible. To alleviate this problem, several studies~\cite{D3human,sim2025persona} adopt 2D vision foundation models~\cite{sapiens,david} to improve the fidelity of geometry.

% Despite these advances, a critical limitation persists in that existing methods primarily rely on photometric reconstruction loss to supervise semi-rigid deformations. This supervision is fundamentally ill-posed under monocular observations due to appearance-geometry ambiguities, such as textureless and self-occluded regions. Such ambiguities frequently trap optimization in local minima, leading to implausible geometry or temporally inconsistent deformations. To address this, recent works have explored temporal correspondences across frames. 

From the perspective of dynamic object reconstruction~\cite{Newcombe_2015_CVPR,Jafarian_2021_CVPR_TikTok}, which prioritizes underlying geometry recovery rather than synthesizing novel view images, moving objects can be reconstructed if dense correspondences can be found between consecutive frames. These correspondences can be parameterized as 6-DoF rigid transformations~\cite{Jafarian_2021_CVPR_TikTok,lei2025mosca}, optical flow between images~\cite{teed2020raft,wang2024sea}, or can be implicitly embedded into learnable motion bases~\cite{kratimenos2024dynmf,wang2025shape}. 
% In other words, understanding temporal correspondences it the key to recovering a moving object. 
While effective for general dynamic scenes, these strategies remain suboptimal for avatar reconstruction because of frequent self-occlusions and non-rigid deformations. 

% Several studies find 6-DoF rigid transforms~\cite{Jafarian_2021_CVPR_TikTok,lei2025mosca} or train learnable motion bases~\cite{kratimenos2024dynmf,wang2025shape}. In the similar context, employing pre-trained optical flow networks~\cite{teed2020raft,wang2024sea} is a popular approach 

% data-driven priors by using pre-trained optical flow networks~\cite{teed2020raft,wang2024sea} to compute reprojection errors between adjacent frames, thereby enforcing temporal smoothness and geometric coherence. While effective for general dynamic scenes, these strategies remain suboptimal for avatar reconstruction. 
% Reconstructed dynamic objects are typically non-animatable and fail to provide plausible completion of unseen regions, both essential for avatars. 
% Moreover, rigid or low-DoF motion constraints are overly restrictive, as LBS efficiently captures the majority of human articulation. 
% Finally, general-purpose data-driven priors, including optical flow and 2D tracking, underperform compared to human-specific priors~\cite{sapiens,david} and struggle to track complex non-rigid human motion~\cite{kim2025learning}.

In this paper, we propose neural avatar representation named as JacobianAvatar providing semi-rigid deformations of a human body, plausible rendering, and temporal consistency, from a single monocular video. The core of JacobianAvatar is the use of hierarchical neural Jacobian fields (NJF)~\cite{NJF} that encode per-triangle deformation gradients conditioned on body poses. We first train coarse Jacobian fields to learn large and global deformations, and then, find Jacobian fields capture high-frequency details subtle local deformations such as wrinkles, enabling scalable and topology-preserving animation of human body. Furthermore, to effectively address the inherent ambiguities of sparse monocular observations, we introduce three geometric constraints. First, we modify the standard Poisson solver of NJF by incorporating a screening term to update the neural Jacobian fields from partial observations. Second, we propose a signed distance function (SDF)-based Jacobian regularizer, which preserves the global shape of the avatar by regularizing the surface normals of its canonical mesh. Finally, to enforce temporal consistency, we formulate a deformation-guided residual flow loss which minimizes the discrepancy between the projected 2D motions of our 3D vertex flows—derived from the NJF deformations—and dense 2D optical flow estimations~\cite{wang2024sea}. 
By integrating expressive deformation modeling, effective shape constraints, and geometrically grounded temporal regularization, our approach generates fully animatable and temporally coherent avatars with fine-grained dynamic detail from limited input. We experimentally verify the performance of our method against state-of-the-art methods~\cite{vid2avatar,exavatar, facavatar, lsavatar} in geometric fidelity, deformation realism, and perceptual quality.

%% file: sec_camera_ready/2_related.tex
\section{Related Work}\label{sec:related}

There are many approaches to generate animatable human avatars from images or videos. One of the most popular approach is to utilize the template human models~\cite{SMPL:2015, pavlakos2019smplx} as they give general information of articulated motion, shape, and expressions. With the aid of template human models, many studies adopt recent technologies such as neural radiance fields (NeRF), 3D Gaussian Splatting and diffusion networks, to better recover accurate and realistic avatars in motion.

For making an animatable avatar, it is essential to define an avatar in the canonical space, this can be done by inverse warping reconstructed 3D models with pre-defined LBS weights of a template model. SCANimate~\cite{bhatnagar2021scanimate} introduces a weakly supervised framework for learning pose-conditioned clothed avatar networks from raw scans, optimizing an implicit occupancy field of a canonical model. % while predicting per-vertex displacement vectors conditioned on joint angles and body shape parameters.
SCALE~\cite{ma2021scale} represents clothed humans as a hierarchical surface codec comprising articulated local elements, enabling decoupled modeling of body and garment dynamics.  
SNARF~\cite{chen2021snarf} proposes a differentiable forward skinning formulation that bridges neural implicit representations and LBS, allowing gradient-based optimization of non-rigid surface deformation.  

The majority of studies utilize monocular videos to capture both global and local deformations using various neural scene representations. NeuMan~\cite{jiang2022neuman} introduces neural human radiance fields from monocular video with human-centric regularizers. SelfRecon~\cite{jiang2022selfrecon} proposes fully self-supervised training through canonical pose alignment and forward-backward deformation consistency. Vid2Avatar~\cite{vid2avatar} proposes a framework for learning avatars from in-the-wild monocular videos, leveraging implicit neural representation with a ray opacity sparsity regularizers for robust geometric representation. CanonicalFusion~\cite{shin2024canonicalfusion} generates drivable 3D human avatars from multiple images via canonical space fusion, but it limits expressive local deformations and results in overly smooth clothing details. More recently, some works~\cite{lsavatar,facavatar} specifically target semi-rigid deformations. LSAvatar~\cite{lsavatar} represents the avatar through pose conditioning using graph neural networks to encode locality-sensitive deformation. FacAvatar~\cite{facavatar} decouples deformation into coarse and fine deformations by controlling the frequencies of positional encodings. However, being built on NeRF architectures, they inevitably suffer from slow rendering and high-frequency artifacts stemming from tightly coupled geometry and texture representations.

%by enforcing monotonic depth ordering and surface smoothness priors, 
%Vid2Avatar~\cite{vid2avatar} proposed a framework for learning animatable avatars from in-the-wild monocular videos, leveraging implicit signed distance fields (SDF) and sparsness regularizers of ray opacity for robust and accurate geometric representation. %with learned 3D deformation grids and self-supervised canonicalization via cycle-consistent density rendering and photometric reconstruction losses. 

Moreover, a growing number of studies employ 3D Gaussian Splatting (3DGS) to achieve real-time rendering speeds while maintaining high visual quality for human avatars.
ExAvatar~\cite{exavatar} extends this paradigm to expressive whole-body reconstruction by anchoring 3D Gaussians to a canonical human template mesh. %and applying localized deformation MLPs to model fine-grained articulation of hands, face, and clothing.  
PERSONA~\cite{sim2025persona} generalizes the framework to generated monocular videos, optimizing identity-preserving canonical geometry from a single reference image. %via cross-frame feature correspondence and deformation regularization.  
Vid2Avatar-Pro~\cite{guo2025vid2avatarpro} reconstructs avatars by leveraging universal priors, achieved through the distillation of human motion and shape information from a large-scale capture dataset. %, combined with ambiguity-aware density modulation and uncertainty-guided optimization.
GaussianAvatar~\cite{hu2024gaussianavatar} models dynamic clothed humans using animatable 3D Gaussians, integrating forward skinning with learned pose-dependent deformation fields to capture non-rigid garment motion.  
Animatable Gaussians~\cite{wang2024animatable} learns a compact pose-conditioned Gaussian attribute map of the avatar via a lightweight convolutional neural networks, enabling the efficient modeling of dynamic 3DGS parameters.
GomAvatar~\cite{Gomavatar} and SplattingAvatar~\cite{guo2024splattingavatar} represent avatars by anchoring 3D Gaussians to a mesh, achieving real-time rendering while preserving high-frequency details of the avatar.
Constrained by the fixed topology and intrinsic rigidity of LBS, these methods may struggle to capture complex non-rigid dynamics like wrinkles and highly deformable garments. % without garment-aware topological adaptation.

On the other hand, diffusion models provide powerful generative priors to resolve geometric and textural ambiguity in occluded regions.  
HumanGaussian~\cite{liu2024humangaussian} generates text-driven 3D clothed humans via 3D Gaussian splatting optimized using score distillation sampling (SDS) from pretrained 2D diffusion models.  
DreamWaltz~\cite{huang2023dreamwaltz} synthesizes complex animatable avatar sequences by combining diffusion-guided motion trajectory sampling with static geometry reconstruction, though dynamic deformation remains limited.  
PSHuman~\cite{li2025pshuman} integrates cross-scale multiview diffusion with SMPL-X body conditioning and explicit remeshing to enforce anatomical consistency, resolve self-occlusions, and produce watertight, animation-ready meshes.  
SiTH~\cite{ho2024sith} employs image-conditioned diffusion with learned skinned body priors to hallucinate photorealistic textures and normal maps in unseen viewpoints.  
HumanRef~\cite{zhang2024humanref} introduces reference-guided diffusion with spatially adaptive region-aware attention to preserve identity-specific details during texture synthesis.  
ReconFusion~\cite{wu2024recon} regularizes sparse-view NeRF reconstruction using 3D-aware diffusion priors, enabling robust novel-view synthesis and geometry completion.  

Despite substantial progress, existing methods suffer from inherent difficulties. Parametric modeling is limited by template topology and LBS rigidity, failing to represent complex semi-rigid deformations. We address these challenges by introducing a monocular framework that learns fine-grained local non-rigid deformations from sparse observations. 
Our approach integrates residual flow fields to enforce sub-pixel temporal consistency beyond photometric losses and regularizes deformation using learned neural Jacobian fields with screened Poisson solver that preserve local area and orientation, enabling robust, topologically flexible reconstruction of both observed and unobserved geometry from monocular video.

%% file: sec_camera_ready/3_proposed.tex
\section{Proposed Method}
\label{sec:proposed}
% 1. 문제의 조건 적기. Monocular video, Camera parameters, and SMPL-X parameters are known.
% Too long 
% (Original) We introduce JacobianAvatar, a self-supervised framework to generate a semi-rigid avatar from a monocular video. The key features are hybrid mesh-Gaussian representation, NJF-based deformation, and a loss function for encouraging temporal consistency. Our avatar represent the deformation on a mesh surface and the fine-detail appearance through 3DGS, which is anchored on the mesh. Furthermore, we propose a novel regularizer that leverages human priors to enforce temporal consistency. This regularizer guides human deformation, and its supervisory signal for the deformation is progressively refined as training progresses, providing increasingly accurate guidance for optimization. The overview of our pipeline is shown in Fig.~\ref{fig:pipeline}. 
% (Original)% How to generate an initial avatar -> take avtantage from exising methods! 
% \subsection{Canonical Avatar Reconstruction}
% As the first step, we define a canonical avatar, \ie, a human model in the resting pose, that can be animated with body pose parameters. The canonical avatar can be either a template human model~\cite{SMPL-X:2019} or a mesh model generated by existing 3D generation pipelines~\cite{xiang2024trellis, hunyuan3d22025tencent}. In the latter case, we transfer the linear blend skinning (LBS) of SMPL~\cite{SMPL:2015} through a robust skin weights transfer~\cite{abdrashitov2023robust}. A vertex can be animated \wrt the change of body pose $\theta$, as long as it has skinning weighs,  

\begin{figure}[pt]
  \centering 
   \includegraphics[width=\linewidth]{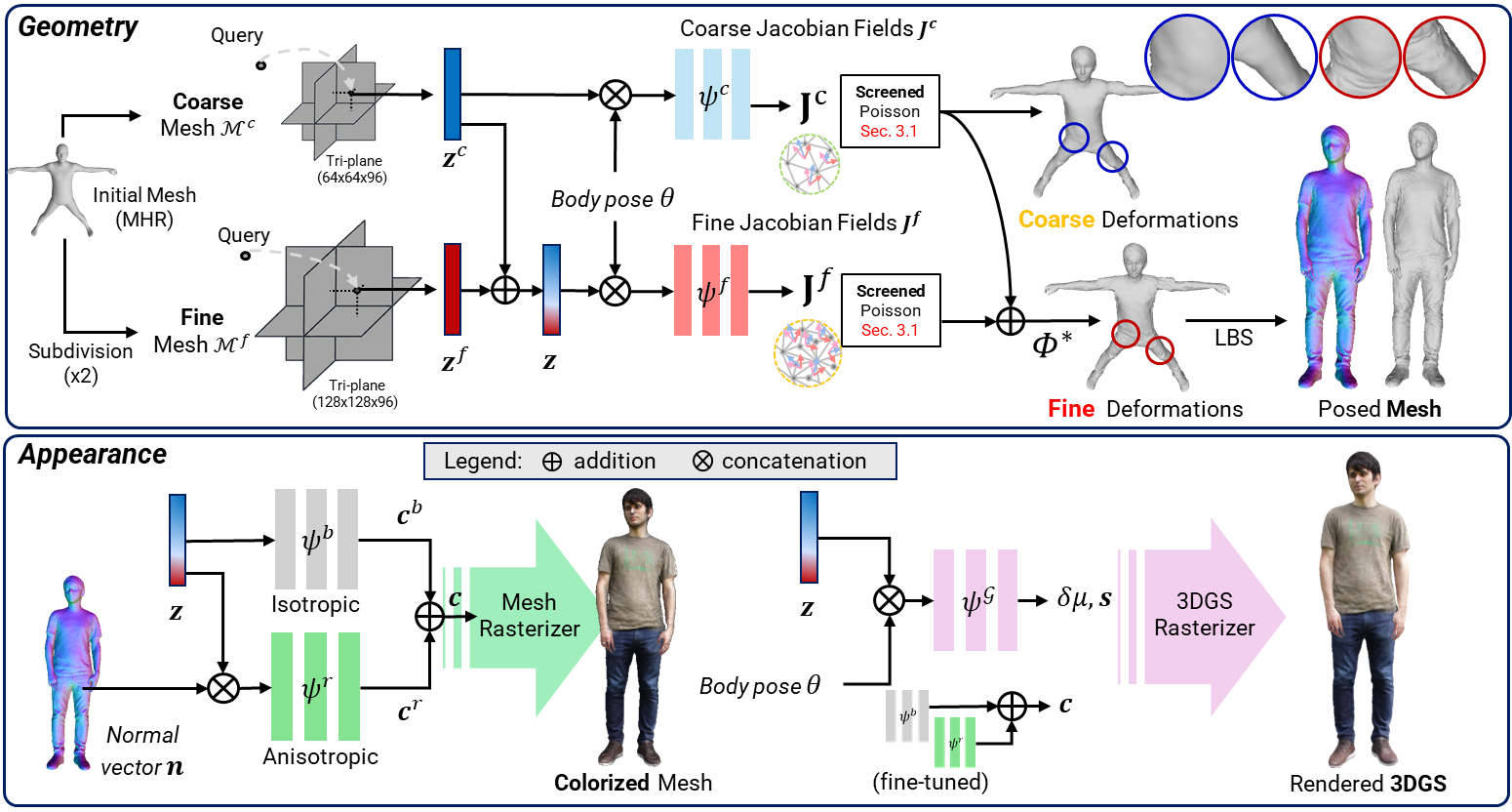}
   \caption{Overview of our pipeline. We first initialize our canonical avatar using a human template mesh and refine it through mesh optimization. Next, we model semi-rigid deformations using two Jacobian fields integrated with a screened Poisson solver, which separately capture deformations in a coarse-to-fine manner. For appearance, we model the mesh textures as normal-conditioned colors. Finally, we incorporate 3DGS anchored to the mesh faces for a more photorealistic rendering of our avatar.}
   %Yellow arrows indicate semi-rigid deformations on the avatar's surface. The MLP $\psi^c$ outputs the pose-independent Jacobian field $\mathbf{J}^c$, which defines the avatar's coarse shape. The MLP $\psi^p$ provides the pose-dependent Jacobian field $\mathbf{J}^p$, which generates local details.
   %After deformation prediction, we compute the 3DGS parameters via the MLP $\psi^\mathcal{G}$, to render the final avatar.} %Yellow arrow indicates semi-rigid deformation on the avatars' surface. From J^c, we can reconstruct pose indepdent deformation which reconstruct avatar's coarse shapes. J^p provides pose depdent deformation that generates local deformations. With reconstructed canonical mesh, we render this leveraging LBS, 3DGS parameters and its rasterizer.  
   \label{fig:pipeline}
\end{figure}

% We introduce JacobianAvatar, a self-supervised framework for generating a semi-rigid avatar from a monocular video. 
% Our approach combines a hybrid mesh-Gaussian representation, neural Jacobian fields (NJF)-based deformation, and a temporal consistency loss. The avatar captures surface deformations on a canonical mesh using NJF and represents fine-grained appearance via 3D Gaussians (3DGS) anchored to mesh vertices. To enforce temporal consistency, we propose a novel residual flow regularizer that leverages human priors. This regularizer progressively refines deformation supervision during training, yielding increasingly accurate geometric alignment. 

We present JacobianAvatar that predicts neural Jacobian fields capturing semi-rigid deformations governed by body pose. Initially, we define a canonical mesh, a human model in a T-pose or Da-pose~\cite{jiang2022neuman}, that can be animated using body pose parameters. We initialize the canonical mesh using the momentum human rigs (MHR)~\cite{MHR:2025} template mesh and existing 3D generation pipelines~\cite{xiang2024trellis, hunyuan3d22025tencent} can be also employed. Prior to training neural networks, we update the canonical mesh and body poses via differentiable rendering and mesh optimization~\cite{nicolet2021large} to reduce the gap between the avatar and the template model. Afterward, we train two neural networks that predict Jacobian fields. To this end, we suggest a constrained Poisson solver with signed distance-based regularizer and residual flow-based loss function. Finally, we train an additional network for predicting Gaussians attached to each face. Therefore, the avatar can be animated and rendered in real-time. The overall procedure is illustrated in Fig.~\ref{fig:pipeline}. 

\noindent\textbf{Pre-processing of Canonical Mesh and Body Poses. }
% % (original)%
% Because of an imperfect initial canonical mesh and body poses, we apply following procedures to refine them. First, we apply continuous remeshing~\cite{palfinger2022continuous} as this step is crucial for applying differential rendering~\cite{Laine2020diffrast}. Then, we update the vertices and body poses for the entire video frames via the differential rendering. More specifically, we predict normal maps and masks from Sapiens~\cite{sapiens} and compare them to the rendered normal maps and masks to update the vertex positions of the initial mesh. Lastly, we subdivide the mesh to capture high-fidelity details from the next steps. 
% Since the initial canonical mesh, \ie, MHR~\cite{mhr}, has different 
Due to imperfections of the initial canonical mesh and body poses, we refine them by using a differentiable mesh renderer~\cite{Laine2020diffrast}. In this step, we jointly optimize the vertex positions and body poses across all video frames, guided by normal maps and segmentation masks predicted by Sapiens~\cite{sapiens}. Formally, each vertex in the canonical mesh is transformed by LBS weights, 
\begin{equation}
  \label{eq:LBS}
    V_i^\theta = LBS(V_i^c, \theta) = \sum_{j=1}^{K} w_{i,j} G_j(\theta) V_i^c, 
\end{equation}
where $V_i^c$ is the position of the $i$-th vertex in the canonical avatar, $\theta$ is the body pose of ${K}$ joints, $V_i^\theta$ is the deformed $V_i^c$ with $\theta$, $G_j(\theta) \in \mathbb{R}^{4 \times 4}$ is the bone transformation matrix of the $j$-th joint derived from $\theta$, $w_{i,j}$ is the skinning weight of $i$-th vertex with respect to the $j$-th joint. Then, we minimize errors between rendered and predicted normal maps and masks. 
After the convergence, we define the updated canonical mesh as $\mathcal{M}^c$, and subdivide the same mesh twice, denoted as $\mathcal{M}^f$. Although we used MHR as a template model, LBS weights are adopted from SMPL-X~\cite{SMPL-X:2019} which are propagated using robust skin weight inpainting~\cite{abdrashitov2023robust}. 

\subsection{Semi-rigid Deformation with Neural Jacobian Fields} %screened poisson 으로 수정. poisson solver -> constrained poisson 설명 - 늘어나는 거 visualize. (toy example/ablation study)
%

% SDF regularizer + OF loss .
Let the canonical triangle mesh obtained after the pre-processing step as $\mathcal{M} = \{\mathcal{V}, \mathcal{F}\}$, where $\mathcal{V}\in \mathbb{R}^{N \times 3}$ and $\mathcal{F} \in \mathbb{Z}^{M \times 3}$ are vertex positions and faces. % and vertex colors, respectively. 
Following the NJF~\cite{NJF}, the mapping function $\Phi$ can transform $\mathcal{V}$ to the new positions, which is defined as
\begin{equation}
    \label{eq:jacobian_poisson}
    \Phi^* = \min_{\Phi} \sum_{a_i \in \mathcal{F}} |a_i|  \|\nabla_i(\Phi) - J_i \|_2^2, 
\end{equation}
where $\nabla_i$ denotes the gradient operator for each face. $|a_i|$ and $J_i$ are the area and a Jacobian matrix of $i^\mathrm{th}$ face. Jacobian field $\mathbf{J}$ is a set of Jacobian matrices that deforms the canonical mesh. Given $\mathbf{J}$, the optimal deformation map $\Phi^*$ can be obtained by solving the Poisson equation~\cite{NJF}. Here, we add a screening term~\cite{kazhdan2013screened} to the objective function, 
\begin{equation}
    \label{eq:screend_jacobian_poisson}
    \Phi^* = \min_{\Phi} \left( \sum_{a_i \in \mathcal{F}} |a_i| \|\nabla_i(\Phi) - J_i \|_2^2 + \lambda_{\Phi}\|\Phi -  \mathcal{V} \|_2^2 \right),  
\end{equation}
where $\lambda_{\Phi}$ is a balancing parameter. The screening term tries to keep the positions of vertices, which is required for invisible and occluded surfaces. In monocular settings, as a result, the screened Poisson behave more reliably compared to the standard Poisson. 
The solution to Eq.~\eqref{eq:screend_jacobian_poisson} can be obtained by solving the linear system, 
% Eq.~\eqref{eq:screend_jacobian_poisson} is can be solved effectively by differentiable Cholesky decomposition~\cite{},
\begin{equation}
    \label{eq:screened_poisson_solution}
     (L+\lambda_{\Phi} I) \Phi^* =\mathcal{A} \nabla^T \mathbf{J} + \lambda_{\Phi}\mathcal{V},
\end{equation}
where $L$ is cotangent Laplacian matrix, $I$ is identity matrix, and $\mathcal{A}$ is the mass matrix of $\mathcal{M}$. The optimal mapping function can be solved through differentiable Cholesky decomposition. In the following section, we explain how we train neural networks to predict optimal Jacobian fields. 

\subsection{Self-supervised Training of Neural Jacobian Fields}
\label{sec:training}
% Terminology 정의 
We train five MLP networks, $\psi^{c}$, $\psi^{f}$, $\psi^{b}$, $\psi^{r}$, $\psi^{g}$ that predict coarse and fine Jacobian fields, colors, and remaining Gaussian properties. First of all, $\psi^{c}$ and $\psi^{f}$ are trained to capture both coarse and fine deformations from a video observation. $\psi^{c}$ predicts a Jacobian matrix for each face at coarse level, 
\begin{equation}
\label{eq:MLP_networks1}
    \mathbf{z}^c_i = \mathcal{E}^c(x_i^c),\ J_i^c = \psi^{c}(\mathbf{z}^c_i, \theta), 
\end{equation}
where $\mathcal{E}^c(x_i^c)$ extracts a spatial latent code from a tri-plane, and $\theta$ is the body pose parameters. In this step, $\psi^{c}$ is trained \wrt the coarse mesh $\mathcal{M}^c$. Afterward, we train $\psi^{f}$ to capture fine details, 
\begin{equation}
\label{eq:MLP_networks}
    \mathbf{z}^f_i = \mathcal{E}^f(x_j^f),\ J_i^f = \psi^{f}(\mathbf{z}^f_i + \mathbf{z}^c_i, \theta). 
\end{equation}
Here, the network structure is same as the coarse MLP but the coarse feature $\mathbf{z}^c_i$ is added to the output of $\mathcal{E}^f(x_j^f)$. Since we train $\psi^{f}$ for a subdivided mesh $\mathcal{M}^f$ with the aid of coarse latent features, this network capture both coarse and fine deformations. 
For training the appearance of an avatar, we train two MLP networks,  
\begin{equation}
    \label{eq:color network}
     \mathbf{c}^b_i=\psi^{b}(\mathbf{z}_i),  \  \mathbf{c}^r_i=\psi^{r}(\mathbf{z}_i, \mathbf{n}_i), \ 
     \mathbf{c}_i =  \mathbf{c}^b_i + \mathbf{c}^r_i, 
\end{equation}
where $\psi^{b}$ and $\psi^{r}$ are shallow MLP networks to predict isotropic and anisotropic components of colors. We adopt this idea from ExAvatar~\cite{exavatar}, instead of using spherical harmonics. Finally, the last network $\psi^{g}$ predicts scale and offset parameters of Gaussians attached to each face.

Since our representation is a hybrid of mesh and 3DGS, we use both mesh renderer $\mathcal{R}_\mathrm{mesh}$ and a GS rasterizer $\mathcal{R}_\mathrm{gs}$. We first update the geometry and color with $\mathcal{R}_\mathrm{mesh}$, 
\begin{equation}
\label{eq:mesh_renderer}
\begin{split}
     \mathbf{\hat{I}}^t, \mathbf{\hat{N}}^t, \mathbf{\hat{A}}^t =  
     \mathcal{R}_{\mathrm{mesh}}( LBS( \Phi_{\theta_t} (\mathcal{M}), \theta_t), P_t),
 \end{split}
\end{equation}
where the mesh rasterizer renders an image $\mathbf{\hat{I}}^t$, a normal map $\mathbf{\hat{N}}^t$, and a foreground mask $\mathbf{\hat{A}}^t$ of a posed mesh. The mapping functions $\Phi_\theta$ can be calculated by solving Eq.~\eqref{eq:screened_poisson_solution} that gives a deformed mesh. Here, $t$ indicates the time stamp for the body pose $\theta_t$ and camera projection matrix $P_t$ because we handle a video sequence. The rendered outputs are compared to self-supervision signals using Sapiens~\cite{sapiens}. We train ($\psi^c$, $\mathcal{E}^c$), ($\psi^f$, $\mathcal{E}^f$), and ($\psi^b$, $\psi^r$) sequentially. Afterward, we use $\mathcal{R}_\mathrm{gs}$ to update Gaussians to recover high-fidelity appearance. We define a set of 3D Gaussians $\mathcal{G}$ as
\begin{equation}
  \mathcal{G}_i = \{\bm{\mu}_i, \delta \bm{\mu}_i, \mathbf{s}_i, \mathbf{q}_i, o_i, \mathbf{c}_i\},
\end{equation}
where $\bm{\mu}_i$ is the face center after deformation, \ie, $\Phi_\theta (x_i)$, $\delta \bm{\mu}_i$ is an offset vector, $\mathbf{s}_i$ is scale, $\mathbf{q}_i$ is quaternion, $o_i$ is opacity, and $\mathbf{c}^r_i$ is the face color calculated in Eq~\eqref{eq:color network}. We predict an offset and a scale as, 
\begin{equation}
    \label{eq:GS_network}
    \delta \bm{\mu}_i, \mathbf{s}_i  =  \psi^\mathcal{G}(\mathbf{z}_i,\theta)
\end{equation}
where $ \psi^\mathcal{G}$ is an MLP network and $\mathbf{z}_i$ is from Eq.~\eqref{eq:MLP_networks}.
Finally, the network is updated by minimizing the input image with the rendered image, 
\begin{equation}
\label{eq:gs_renderer}
\begin{split}
     \mathbf{\hat{I}}^t =  
     \mathcal{R}_{\mathrm{gs}}( LBS(\mathcal{G}, \theta_t), P_t),
 \end{split}
\end{equation}
where $LBS(\cdot)$ warps 3D Gaussians to the body pose of $t^\mathrm{th}$ frame. 
\subsection{Training Objectives}

We propose two total loss functions for the mesh renderer and GS renderer. The first loss function is defined as, 
\begin{equation}
\label{eq:level_loss}
\begin{split}
\mathcal{L}_{\text{mesh}} =& \lambda_{\text{L1}} \mathcal{L}_{\text{L1}}(\mathbf{\hat{I}}^{t}, \mathbf{I}^t) 
    + \lambda_{\text{lpips}} \mathcal{L}_{\text{lpips}}(\mathbf{\hat{I}}^{t}, \mathbf{I}^t) 
    + \lambda_{\text{ssim}} \mathcal{L}_{\text{ssim}} (\mathbf{\hat{I}}^{t}, \mathbf{I}^t) \\ 
    &+ \lambda_{\text{mask}} \mathcal{L}_{\text{mask}}(\mathbf{\hat{A}}^{t}, \mathbf{A}^t) 
    + \lambda_{\text{normal}} \mathcal{L}_{\text{normal}}(\mathbf{\hat{N}}^{t}, \mathbf{N}^t) \\
    &+ \lambda_{\text{residual}} \mathcal{L}_{\text{residual}}(\mathbf{I}^t, \mathbf{I}^{t-1}, \mathbf{W}^{t}) 
    + \lambda_{\text{SDF}} \mathcal{L}_{\text{SDF}}(\Phi_{\theta_t}(\mathcal{M})) \\
    &+ \lambda_{\text{reg}} \mathcal{L}_{\text{reg}}(\mathbf{J}_{\theta_t}) 
    + \lambda_{\text{lap}_{\mathcal{V}}} \mathcal{L}_{\text{lap}}(\Phi_{\theta_t}(\mathcal{V}), \mathcal{F}) \\
    &+ \lambda_{\mathbf{C}^b} \mathcal{L}_{\text{dual-lap}}(\mathbf{C}^{b}, \mathcal{F}) 
    + \lambda_{\mathbf{C}} \mathcal{L}_{\text{dual-lap}}(\mathbf{C}, \mathcal{F})
\end{split}
\end{equation}
where $L_\mathrm{L1}$, $L_\mathrm{lpips}$, and $L_\mathrm{ssim}$ compute L1, perceptual, and structural similarity losses between input and rendered images. $L_\mathrm{normal}$ is a normal cosine similarity loss with a pseudo ground truth normal map from a pretrained model~\cite{sapiens}. $L_\mathrm{mask}$ is an L1 loss between a rendered foreground mask and a pseudo ground truth mask from a pretrained model~\cite{sapiens}. $L_\mathrm{reg}$ is an L1 regularizer on the pose dependent deformation maps derived from the $\mathbf{J}_{\theta_t}^{c}$ and $\mathbf{J}_{\theta_t}^{f}$ by solving the screened Poisson equation~\eqref{eq:screened_poisson_solution}. $\mathcal{L}_{\text{lap}}$ is a Laplacian regularizer and $\mathcal{L}_{\text{dual-lap}}$ is a dual Laplacian regularizer. For color, we regularize for both anisotropic and combined colors, $\mathbf{C}^{b}$ and $\mathbf{C}$. Note that this loss function is applied to coarse and fine meshes in a sequential manner. The second loss function is defined as, 
\begin{equation}
\label{eq:total_loss}
\begin{split}
    \mathcal{L}_{\text{gs}} ={}& \lambda_{\text{L1}} L_{\text{L1}}(\mathbf{\hat{I}}^t, \mathbf{I}^t) + \lambda_{\text{lpips}} L_{\text{lpips}}(\mathbf{\hat{I}}^t, \mathbf{I}^t) 
    +  \lambda_{\text{ssim}} L_{\text{ssim}}(\mathbf{\hat{I}}^t, \mathbf{I}^t) \\
     &+ \lambda_{\text{scale}} L_{\text{scale}}(\mathbf{s})  +\lambda_{\text{offset}} L_{\text{offset}}(\delta \bm{\mu})
     + \lambda_{\text{lap} \mathbf{s}} L_{\text{lap}}(\mathbf{s}) \\
     &+ \lambda_{\text{lap}\delta \bm{\mu}} L_{\text{lap}}(\delta \bm{\mu})  
     +  \lambda_{\text{residual}} L_{\text{residual},}(\mathbf{I}^t, \mathbf{I}^{t-1},  \mathbf{W}^t )
\end{split}
\end{equation}
where $L_{\text{scale}}$ and $L_{\text{offset}}$ are L2 regularizers on scale $\mathbf{s}$ and offsets $\delta \bm{\mu}$, respectively. For temporal consistency, we adapt the residual flow loss on 3D Gaussians. We obtain a 2D flow map~$\mathbf{W}^t$ by assigning a displacement vector as an attribute to each 3D Gaussian and rendering the flow map using alpha-blending.

Here, we propose $\mathcal{L}_\text{residual}$ and $\mathcal{L}_\text{SDF}$ loss functions to the fidelity and temporal consistency of an avatar, which we explain in detail in the following sections. 

\noindent\textbf{Deformation-guided Residual Flow. }
We claim that relying on per-frame observations is insufficient to capture fine-grained local deformations. Therefore, we propose a temporal consistency regularizer to capture temporal changes. First, we select two frames, \eg, at time $t$ and $t-1$, and determine the difference of projected $V_i$ at different time steps. Then, the projected point can be denoted, 
\begin{equation}
\label{eq:residual1}
    \mathbf{u}_i^t = \pi( LBS(\Phi_{\theta_t}(V_i), \theta_t), P_t), 
\end{equation}
where vertex $V_i$ is deformed by the predicted mapping function $\Phi_{\theta_t}$ followed by the LBS function, and $\pi$ is a projection function with the camera projection matrix $P_t$. Then, we further define a 2D displacement vector, 
\begin{equation}
\label{eq:residual2}
    \delta \mathbf{u}_i^t = \mathbf{u}_i^t - \mathbf{u}_i^{t - 1}, 
\end{equation}
which is the displacement of $V_i$ in pixels. Here, we assign $\delta \mathbf{u}_i^t$ as the vertex attribute; therefore, it gives a 2D flow map between the two frames. We define this as $\mathbf{W}^t$ as the 2D flow map between images. Instead of comparing $\mathbf{W}^t$ to the optical flow between the two frames, we find a residual flow by using a pre-trained optical flow network $\mathcal{F}$~\cite{wang2025waft}, 
\begin{equation}
\label{eq:residual3}
    \Delta \mathbf{W}^t = \mathcal{F}(\mathbf{I}^t, \mathbf{I}^{t-1}, \mathbf{W}^t). 
\end{equation}
where $\mathbf{{I}}$ is a ground truth image. Here, $\mathcal{F}$ search correspondence given initial displacement map $\mathbf{W}^t$, therefore, predicted residual $\Delta \mathbf{W}^t$ becomes zero if $\mathbf{W}^t$ is perfect. To this end, we minimize the residual flow, 
\begin{equation}
\label{eq:residual4}
    L_\mathrm{residual} = ||\Delta \mathbf{W}^t||_1, 
\end{equation}
to further constrain the temporal consistency between two frames. 

\noindent\textbf{Signed distance-based Jacobian regularization. }
% After the pre-processing step, we define a Signed Distance Function (SDF) within a pre-defined volume, and compute normal vectors by discretizing SDF values within a small volume. Since predicted Jacobian matrices give noisy deformations, we regularize the direction of Jacobian matrix. At the surface, the gradient of the SDF is parallel but in the opposite direction to the surface normal vector~\cite{meshsdf}. Since it is possible to computer 
Sparse observations and self-occluded regions of the avatar frequently generate unwanted deformations, such as severe mesh stretching and flipped faces. To address this, we propose a SDF-based regularizer to maintain the global shape of the canonical avatar. To achieve this, we set a voxel grid containing the discrete SDF values of the initial coarse mesh $\mathcal{M}^c$. Then, we compute the spatial gradients of signed distance values at the deformed vertices and faces, and set approximate normals as flipped spatial gradients~\cite{remelli2020meshsdf}. 

Surface normals of the initial mesh by calculating the spatial gradients within the SDF voxel grid. Using these derived normals as a geometric prior, we regularize the face and vertex normals of the deformed canonical avatar via a cosine similarity loss, 
\begin{equation}
    \label{eq:sdf_normal_loss}
    \mathcal{L}_{\text{SDF}} = \lambda_{\text{face}} \sum_{a_i \in \mathcal{F}} \left( 1 - \hat{\mathbf{n}}^a_i \cdot {\mathbf{n}}_{\text{SDF}}(x_i) \right) + \lambda_{\text{vertex}} \sum_{v_j \in \mathcal{V}} \left( 1 - \hat{\mathbf{n}}^v_j \cdot {\mathbf{n}}_{\text{SDF}}(v_j) \right),
\end{equation}
where $x_i$ is the center of the $i$-th face of the mesh, and $\mathbf{n}^a_i$ and $\mathbf{n}^v_j$ denote the face and vertex normal vectors of the deformed canonical avatar, respectively. Furthermore, $\hat{\mathbf{n}}_{\text{SDF}}(x)$ is the normal extracted from the SDF voxel grid by normalizing the spatial gradient of the SDF volume at position $x$. Through these geometric priors, the Jacobian matrices evolve to neighboring iso-surfaces rather than abruptly changing positions. 

%% file: sec_camera_ready/4_experiment.tex
\section{Experimental Results}
\label{sec:exp}
\subsection{Implementation details} 
%We implement
%A6000 ada Gpu
 
We initialize our mesh using the MHR~\cite{MHR:2025} template at Level of Detail (LoD) 3. For the screened Poisson solver, we empirically set $\lambda_{\Phi}$ to 5 and 1 for the coarse and fine meshes, respectively. The tri-plane resolutions are set to 64 for $\mathcal{E}^c$ and 128 for $\mathcal{E}^f$. Each plane has a feature dimension of 32, and we concatenate the features extracted from all three planes. For stable training, following NJF~\cite{NJF}, we also initialize the last layer of both MLPs, $\psi^c$ and $\psi^{f}$, with zero weights and biases. We train our avatar on a single RTX 6000 Ada GPU. Including preprocessing, the training process takes about 10 hours, depending on the video length.

%WeFor detials of weights of losses are 
% For hyper-parameters, we set $\lambda_{L1}$, $\lambda_{lpips}$, $\lambda_{ssim}$, $\lambda_{mask}$, $\lambda_{residual}$, $\lambda_{reg}$, $\lambda_{laplacian}$, $\lambda_{scale}$, and $\mathcal{L}_{\text{lap}}$,to 10.0, 2.0, 2.0, 1.0, 50.0, 1.0, 1000.0, 10.0, and 0.1, respectively.

%We implement all MLP networks using FINER~\cite{liu2024finer} to effectively represent high-frequency details. $\psi^c$ and $\psi^p$ are three layer MLPs. The network $\psi^\mathcal{G}$ is composed of four distinct MLPs, each outputting a specific parameter of the 3DGS $\mathcal{G}$. We train our avatar with single A6000 ada GPU, training takes approximately 8 hours but it depends on the video length. For hyper-parameters, we set $\lambda_{L1}$, $\lambda_{lpips}$, $\lambda_{ssim}$, $\lambda_{mask}$, $\lambda_{residual}$, $\lambda_{reg}$, $\lambda_{laplacian}$, $\lambda_{scale}$, and $\lambda_{offset}$ to 10.0, 2.0, 2.0, 0.1, 100.0, 1.0, 1000.0, 10.0, and 0.1, respectively.

\subsection{Datasets} 

We evaluate our method on four public datasets, which include diverse identities, poses, and environments.

\noindent\textbf{MonoPerfCap dataset~\cite{xu2018monoperfcap}.} The MonoPerfCap dataset is captured in in-the-wild environments, including indoors and outdoors, and contains natural human poses. It is recorded under stable lighting conditions, and subjects typically wear non-loose clothing, making it ideal for evaluating the reconstruction of semi-rigid deformations. This dataset also provides high frame rate videos, yielding temporally dense and continuous motion. Following Vid2Avatar-Pro~\cite{guo2025vid2avatarpro}, we split the dataset into training and testing sets, comprising the first $80\%$ of frames and the remaining frames, respectively. This split allows us to evaluate generalization performance to novel views.

\noindent\textbf{SynWild dataset~\cite{vid2avatar}.} The SynWild dataset is a synthetic dataset that provides ground-truth meshes, enabling the quantitative evaluation of geometric accuracy. It contains clothing deformations, making it particularly suitable for assessing semi-rigid deformations. Since this dataset lacks camera parameters, we apply Iterative Closest Point (ICP) registration to align reconstructed avatars with the ground-truth meshes, following the evaluation protocol of the previous work~\cite{vid2avatar}.

\noindent\textbf{NeuMan dataset~\cite{jiang2022neuman}.} The NeuMan dataset consists of outdoor videos including subjects performing repetitive motions. Compared to MonoPerfCap dataset, the motions are simple and limited, and cloth deformation is less. Besides, the humans move relatively longer distance than other datasets, leading to lighting variations depending on the position. We follow the official data split, which utilize interval views for testing, making it suitable for evaluating novel view interpolation performance.

\noindent\textbf{DNA-Rendering dataset~\cite{cheng2023dna}} The DNA-Rendering dataset is captured in a multi-view studio with various types of clothing. It contains sequences featuring loose garments with challenging non-rigid deformations. By randomly sampling a single camera per frame, we adapt this dataset to a monocular setting. Using this setup, we evaluate novel view synthesis performance and qualitatively compare mesh reconstruction results with state-of-the-art (SOTA) SDF-based methods.

\subsection{Comparisons to State-of-the-art Methods}
\noindent\textbf{Comparison methods.}
To demonstrate the effectiveness of our representation, we compare our method against five methods that achieve SOTA results in geometric accuracy and rendering quality. 
Vid2Avatar~\cite{vid2avatar}, LSAvatar~\cite{lsavatar} and FacAvatar~\cite{facavatar} represent the canonical avatar using a SDF, an approach known for its robustness in capturing geometric details. ExAvatar~\cite{exavatar} and GoMAvatar~\cite{Gomavatar} utilizes 3DGS to produce photorealistic rendering results and achieves robustness in novel pose and novel view synthesis by applying a Laplacian regularizer that is computed using the connectivity of the mesh of the canonical avatar.

%its canonical avatar, shows plausible rendering quality by leveraging regularizers underlying SMPL base mesh.
%우리 represntation의 효과성을 보이기 위해서 geometry accuracy와 rendering quality에서 SOTA 성능을 보이는 두 방법론과 비교하였다. Vid2avatar는 singed distance function을 활용하여 canonical avatar를 표현한 방법으로 geometry detail을 표현하는데 강인한 방법론이다.
%Exavatar는 3DGS로 canonical avatar를 표현한 방법으로 SMPL Base mesh를 활용한 regularizer들을 활용하여 avatar reconstruction에서 SOTA 성능을 보였다. 

\noindent\textbf{Evaluation Protocol.} We quantitatively compare our method with SOTA methods. To assess rendering quality, we measure PSNR, SSIM, and LPIPS~($\times100$). In this evaluation, except for DNA-Rendering dataset, we follow a protocol of previous works, which additionally optimizes the body poses for the test set using rendering losses. This evaluation protocol reduces the dependency of the metrics on the accuracy of the body pose. To evaluate geometric accuracy, we measure the Chamfer Distance (CD) ($\times 1000$), Normal Error (NE) and the F1-score at thresholds of 1 cm and 2 cm against the ground-truth meshes. 
For Normal Error, we render the normals of both the ground-truth and aligned meshes from four orthogonal views (front, back, left, and right), and then measure the cosine similarity in overlapping regions.

\begin{table}[!ht]
\centering
\caption{Quantitative comparison of rendering quality on the
MonoPerfCap, DNA-Rendering, and NeuMan datasets.}
\label{tab:rendering_quant}

%\fontsize{7.5pt}{8.6pt}\selectfont
\setlength{\tabcolsep}{2.0pt}
\renewcommand{\arraystretch}{1.02}
\resizebox{0.85\columnwidth}{!}{%

\begin{tabular}{@{}c>{\raggedright\arraybackslash}p{0.32\columnwidth}@{\hspace{3pt}}ccc@{}}
\toprule
\textbf{Dataset} & \textbf{Method} 
& \textbf{PSNR}$\uparrow$ 
& \textbf{SSIM}$\uparrow$ 
& \textbf{LPIPS}$\downarrow$ \\
\midrule

\multirow{6}{*}{\makecell[c]{\textbf{(a)} MonoPerfCap}}
& GoMAvatar~\cite{Gomavatar}   & 28.35 & 0.975 & \underline{1.95} \\
& Vid2Avatar~\cite{vid2avatar} & 28.74 & 0.977 & 2.06 \\
& LSAvatar~\cite{lsavatar}     & 27.96 & 0.975 & 2.15 \\
& FacAvatar~\cite{facavatar}   & 28.06 & 0.975 & 1.99 \\
& ExAvatar~\cite{exavatar}     & \underline{30.47} & \textbf{0.980} & 2.00 \\
\cmidrule(l){2-5}
& \textbf{Ours}                & \textbf{31.83} & \underline{0.978} & \textbf{1.67} \\

\midrule

\multirow{4}{*}{\makecell[c]{\textbf{(b)} DNA-Rendering}}
& Vid2Avatar~\cite{vid2avatar} & \underline{29.46} & \bf{0.972} & 2.88 \\
& LSAvatar~\cite{lsavatar}     & 27.94 & 0.968 & \underline{2.37} \\
& ExAvatar~\cite{exavatar}     & 28.54 & 0.970 & 2.85 \\
\cmidrule(l){2-5}
& \textbf{Ours}                & \textbf{29.90} & \underline{0.971} & \textbf{2.19} \\

\midrule

\multirow{8}{*}{\makecell[c]{\textbf{(c)} NeuMan}}
& HumanNeRF~\cite{humannerf} & 27.06 & 0.967 & 1.92 \\
& InstantAvatar~\cite{jiang2022instantavatar} & 28.47 & 0.972 & 2.77 \\
& NeuMan~\cite{jiang2022neuman} & 25.48 & 0.966 & 2.87 \\
& Vid2Avatar~\cite{vid2avatar} & 26.87 & 0.969 & 2.41 \\
& 3DGS-Avatar~\cite{qian20233dgsavatar} 
    & \underline{29.75} & \underline{0.975} & \underline{1.75} \\
& GaussianAvatar~\cite{hu2024gaussianavatar} 
    & \textbf{29.94} & \textbf{0.980} & \textbf{1.24} \\
& ExAvatar~\cite{exavatar} 
    & 29.37 & \textbf{0.980} & 1.87 \\
\cmidrule(l){2-5}
& \textbf{Ours} 
    & \underline{29.75} & 0.974 & 1.94 \\

\bottomrule

\end{tabular}
}
\end{table}

\begin{table}[!t]
    \centering
    \caption{
        Quantitative comparison of geometry accuracy on the SynWild dataset.
    }
    \label{tab:synwild_quant}
    \resizebox{0.7\columnwidth}{!}{%
        \begin{tabular}{l l c c c c}
            \toprule
            \textbf{Method} & \textbf{Type} 
            & \textbf{CD} $\downarrow$ 
            & \textbf{NE} $\downarrow$ 
            & \textbf{F1@1cm} $\uparrow$ 
            & \textbf{F1@2cm} $\uparrow$ \\
            \midrule
            GoMAvatar~\cite{Gomavatar}   
                & Explicit & 9.30 & 0.140 & 0.276 & 0.544 \\
            ExAvatar~\cite{exavatar}     
                & Explicit & 9.39 & --    & 0.307 & 0.515 \\
            Vid2Avatar~\cite{vid2avatar} 
                & Implicit & 2.59 & \textbf{0.091} & \underline{0.346} & \underline{0.647} \\
            LSAvatar~\cite{lsavatar}     
                & Implicit & \underline{2.55} & 0.095 & 0.343 & 0.646 \\
            FacAvatar~\cite{facavatar}   
                & Implicit & 2.91 & 0.106 & 0.329 & 0.623 \\
            \midrule
            \textbf{Ours}                
                & Explicit & \textbf{2.46} & \textbf{0.091} & \textbf{0.397} & \textbf{0.681} \\
            \bottomrule
        \end{tabular}
    }
\end{table}

\noindent\textbf{Quantitative results.} 
As shown in Tab.~\ref{tab:rendering_quant} (a), our method outperforms the comparison methods in view extrapolation on the MonoPerfCap dataset. This is attributable to our hybrid mesh-3DGS representation, which provides coherent geometry and high-fidelity appearance in unseen views. Furthermore, the high frame rate of this dataset is beneficial for our deformation guided residual flow loss $L_{residual}$, which in turn contributes to the improved overall rendering performance.

For the DNA-Rendering dataset, we evaluate novel view synthesis results for avatars with loose clothing in Tab.~\ref{tab:rendering_quant} (b). Our pipeline effectively reconstructs these avatars by first optimizing the global shape via an initial mesh optimization step, and subsequently utilizing the network to recover fine-level geometry. Providing more accurate geometry compared to baselines, our method achieves the best rendering quality.

In Tab.~\ref{tab:rendering_quant} (c), we achieve results comparable to SOTA methods on the NeuMan dataset, despite the severe lighting changes in the scenes, which do not explicitly consider. The results for all baseline methods with the exception of ExAvatar, are directly borrowed from previous work~\cite{guo2025vid2avatarpro}. We evaluated ExAvatar using the pretrained weights provided with the author's reimplementation, without rendering the background to ensure a fair comparison with these baselines.

As shown in Tab.~\ref{tab:synwild_quant}, our explicit approach compares favorably against recent implicit methods, achieving the state-of-the-art performance in modeling semi-rigid deformations. Unlike implicit models that struggle to guarantee temporal consistency, we directly deform the mesh via NJF, constraining this explicit deformation through a deformation-guided residual loss. Furthermore, implicit represent may suffer from artifacts during mesh extraction, which limits their ability to preserve high-frequency geometric details.

\begin{figure*}[t!]
  \centering 
   \includegraphics[width=\linewidth]{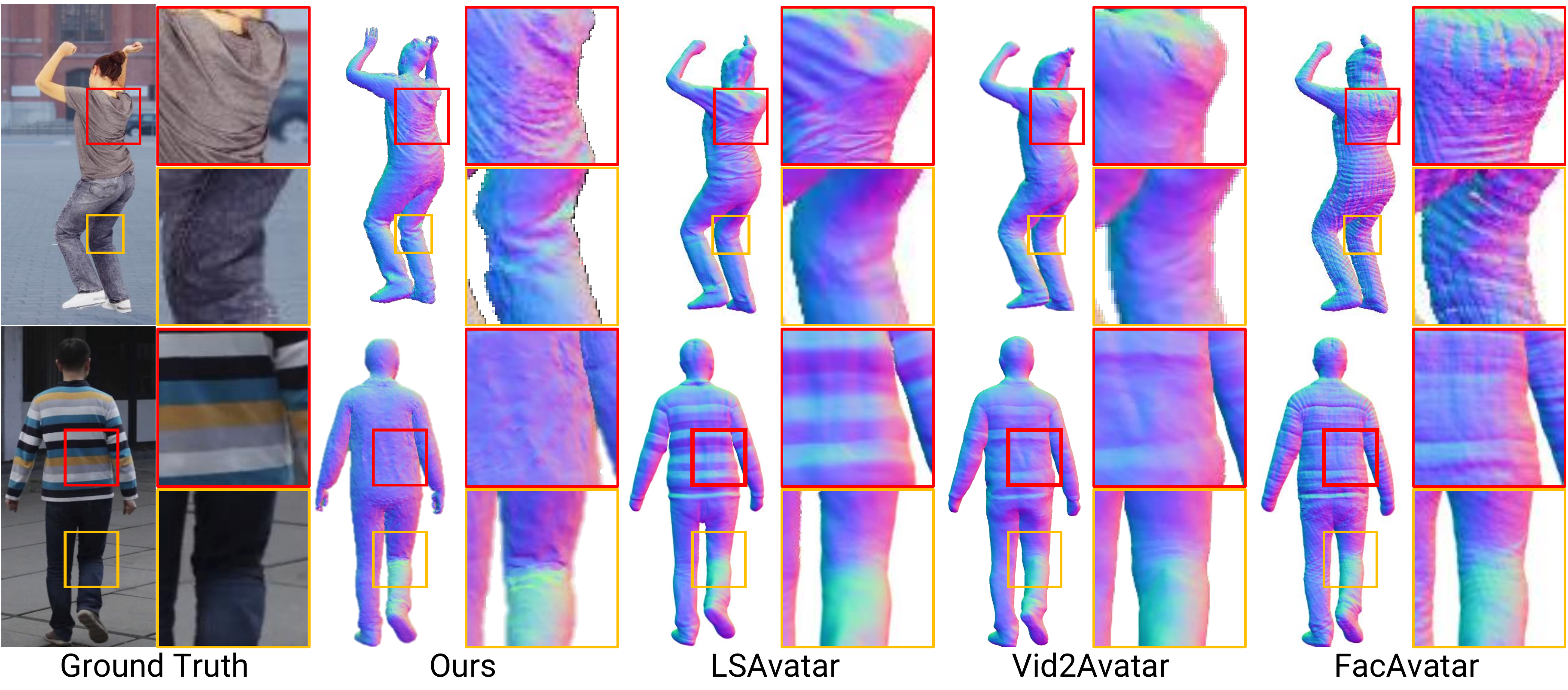}
   \caption{Qualitative comparison of rendered normal maps. Compared methods~\cite{vid2avatar,lsavatar,facavatar} suffer from texture-copying artifacts on the geometry, whereas our method shows smooth surfaces in the textured regions.}
   \label{fig:Synwild geometry qualitative}
\end{figure*}

\begin{figure}[t!]
  \includegraphics[width=\linewidth]{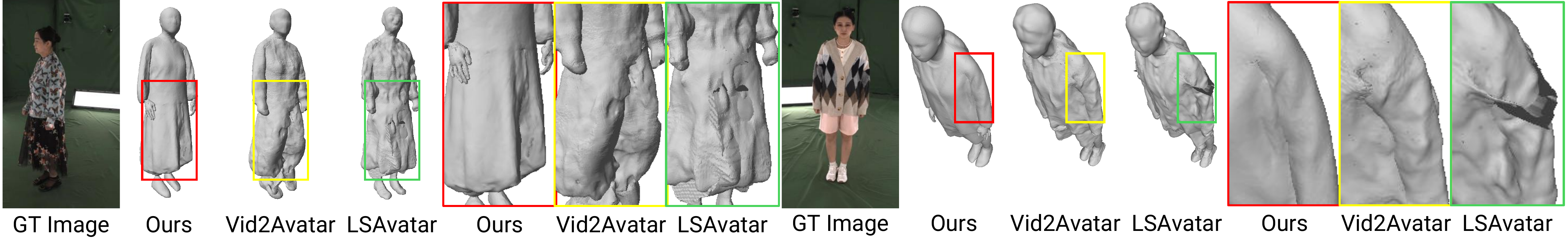}
  \caption{Qualitative comparison of mesh reconstruction with SDF-based methods~\cite{vid2avatar,lsavatar} on the DNA-Rendering dataset.}
  \label{fig:DNA_Rendering_qualitative}
\end{figure}

\begin{figure*}[t!]
  \centering 
   \includegraphics[width=\linewidth]{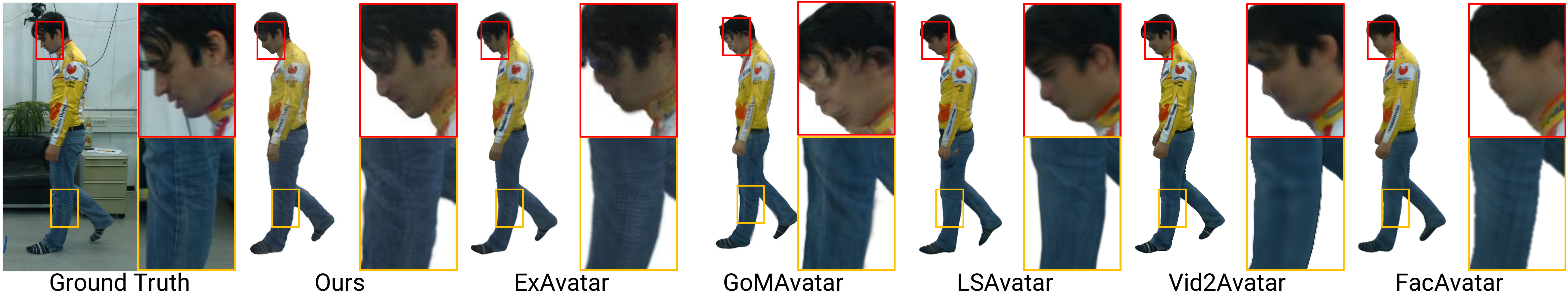}
   \caption{Qualitative comparison of rendering quality with state-of-the-art methods~\cite{exavatar,vid2avatar,Gomavatar,lsavatar,facavatar} on the MonoperfCap Dataset.}
   \label{fig:Monoperfcap qualitative}
\end{figure*}

\noindent\textbf{Qualitative results.} We provide qualitative comparisons from both perspectives: geometry and rendering quality of the avatar. In Fig.~\ref{fig:Synwild geometry qualitative}, we compare rendered normal maps using the Synwild dataset. As shown in the first row, our method recovers the sharp wrinkles in the pants and the T-shirt, with the aid of neural Jacobian fields. More importantly, we observed texture-copying artifact, where 2D textures are incorrectly baked into the 3D geometry when a human wears highly textured cloth (as shown in the bottom row). This indicates geometry and texture are entangled in the existing approach. In our case, we independently capture semi-rigid deformations without optimizing texture-related variables, which alleviated this phenomenon. 

Furthermore, as shown in Fig.~\ref{fig:DNA_Rendering_qualitative}, our method produces cleaner meshes on the DNA-Rendering dataset, Specifically, the proposed pipeline is free from the extraction artifacts inherently caused by the Marching Cubes algorithm, which is essential for mesh reconstruction in previous SDF-based approaches.

%we provide a qualitative comparison of mesh reconstruction results on the DNA-Rendering dataset against previous SDF-based approaches. Unlike these approaches, which rely on marching cubes and often suffer from extraction artifacts, our method is free from such artifacts and produces cleaner meshes.

In Fig.~\ref{fig:Monoperfcap qualitative}, we qualitatively compare rendered images. When the face is seen from the side view, most existing methods show blurry textures, whereas our method shows relatively clean results. This implicitly confirms that the geometry is more accurate compared to SOTA methods. Secondly, in the highlighted pants region, we can observe the proposed method shows sharp and accurate textures. We believe the Jacobian fields are good at capturing local deformations, therefore, attached 3DGS also shows sharp and clean results.

\begin{figure*}[t!]
  \centering 
   \includegraphics[width=\linewidth]{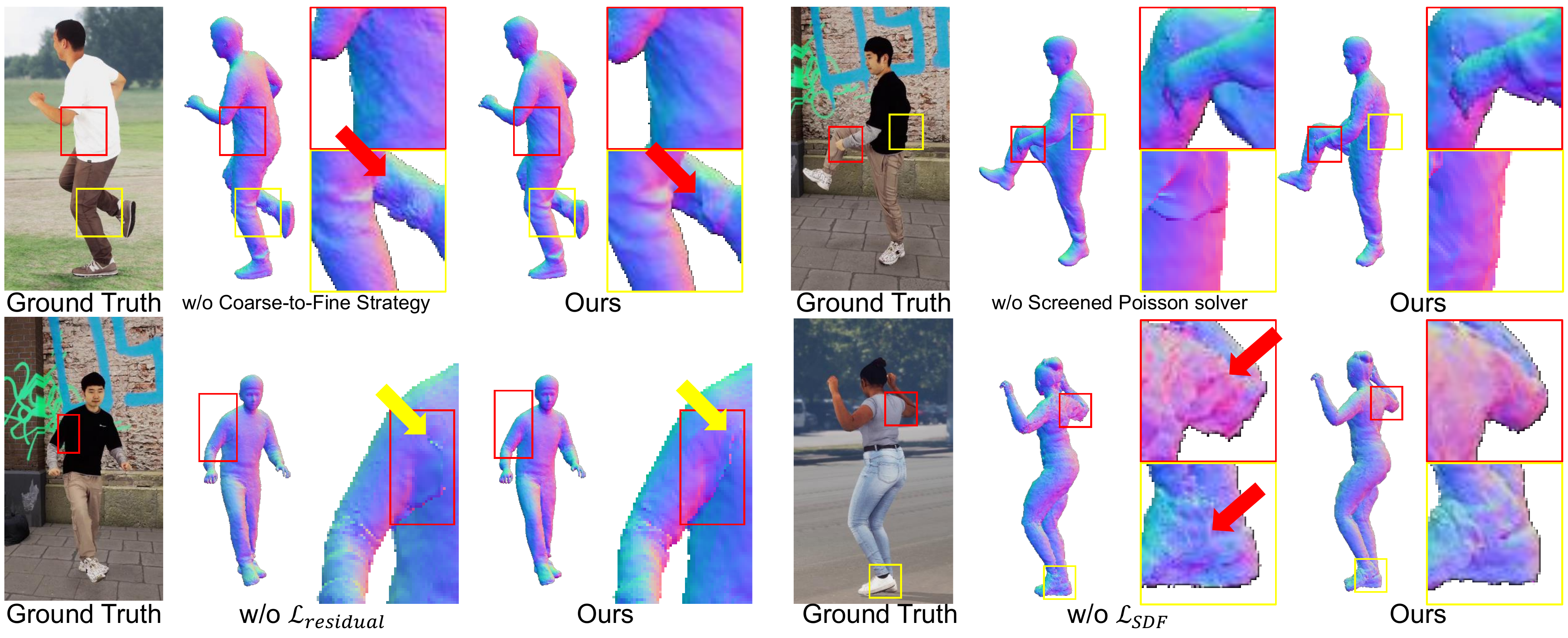}
   \caption{Ablation study on the Synwild dataset. We visualize the effects of each component of our proposed method on the geometry quality by removing them independently.}
   \label{fig:ablation study qualitative}
\end{figure*}

\subsection {Ablation study} %\noindent\textbf{Qualitative results} 
%We demonstrate the role of each component of our model in~\ref{fig:ablation}. 
We analyze the impact of each component of our method qualitatively in Fig.~\ref{fig:ablation study qualitative} and quantitatively in Tab.~\ref{tab:synwild_ablation}.
%In Fig.~\ref{fig:ablation study qualitative} and Tab.~\ref{tab:synwild_ablation}, we analyze the impact of each proposed component in quantitavely and qualitatively. 

\begin{table}[t!]
   % \vspace{-3.5em}
    \centering
    \caption{Ablation studies on the \texttt{00000\_random} scene of the SynWild dataset. We assess geometry quality by removing each component of our method.}
    \label{tab:synwild_ablation} 
    \resizebox{0.6\linewidth}{!}{%
        \begin{tabular}{lcc}
            \toprule
            \textbf{Method} & \textbf{CD($\times1000)$} $\downarrow$ & \textbf{NE} $\downarrow$ \\
            \midrule
            w/o $\mathcal{L}_{\text{SDF}}$           & 3.51  & 0.113 \\
            w/o $\mathcal{L}_{\text{residual}}$      &\underline{3.01}  & \underline{0.106} \\
            w/o Screened Poisson solver                       & 3.28  & \underline{0.106} \\
            w/o Coarse-to-fine Strategy                       & \bf{2.96}  & 0.107 \\

            \midrule
            \textbf{Ours} & $\underline{3.01}$ &  $\mathbf{0.102}$ \\
            \bottomrule
        \end{tabular}%
    }
    %\vspace{-2em}
\end{table}

\noindent\textbf{Coarse-to-fine Strategy.} Optimizing the fine Jacobian field without coarse Jacobian field introduces high-frequency, noisy deformations on the mesh. As shown in the figure, it also fails to capture prominent wrinkles compared to our full model. Therefore, in the table, it performs well in CD, which is relatively insensitive to small surface noise, but performs poorly in NE where these noisy deformations are penalized.

\noindent\textbf{Screened Poisson solver.} Relying solely on the Poisson solver without our formulation causes large stretching artifacts due to the limited observations from monocular videos. However, regularizing the solver using the initial mesh effectively prevents these artifacts. As these large artifacts result in severe geometric distortions, they cause a degradation in the CD. %, as shown in the table.

\noindent\textbf{Deformation-guided Residual Flow Loss ($\mathcal{L}_{residual}$).} This loss is not only helpful for capturing temporal consistency but also plays a crucial role in suppressing artifacts in heavily self-occluded areas like the armpits. Specifically, it mitigates these artifacts by regularizing the noisy flow predictions generated by pretrained models. However, these self-occluded regions occupy only a small portion of the overall body, removing this loss results in remaining a similar CD. However, the resulting local artifacts degrade the surface quality, leading to poor performance in NE. %, as shown in the table.

\noindent\textbf{Signed Distance-based Jacobian Regularization ($\mathcal{L}_{SDF}$).} This term regularizes the global shape of the avatar, effectively preventing artifacts in regions with sparse observations. Consequently, without this regularization, the woman's back, which is a rarely visible area in the video, suffers from severe noisy deformations as shown in the figure. Moreover, omitting this loss makes the model susceptible to overfitting to specific viewpoints, leading to an overall degradation in mesh quality. As a result, the absence of this regularizer causes a performance drop in both CD and NE. %, as reported in the table.

%% file: sec_camera_ready/5_conclusion.tex
\section{Conclusion}
\label{sec:conclusion}

We have proposed a robust framework for generating drivable, semi-rigid avatars from monocular video. By integrating Neural Jacobian Fields (NJF) with a screened Poisson solver, and introducing a deformation-guided residual flow with SDF-based regularization, our method captures semi-rigid deformations while preventing geometric artifacts. We demonstrate effectiveness of our work through comparisons with the-state-of-the-art methods on various scenes.

\noindent\textbf{Limitations and Future Work.} Our method struggles to reconstruct avatars with loose-fitting clothing. However, since it relies on the LBS weights from the template model, \ie, a naked body, loose clothes that behave differently from the inner body cannot be modeled. Next, our current model does not encompass facial expressions. We leave these extensions as future work, where existing facial reconstruction method can be integrated into our framework. 

\noindent\textbf{Acknowledgement}\quad
This work was supported by the Institute of Information \& Communications Technology Planning \& Evaluation (IITP) grants funded by the Korea government (MSIT) (No.RS-2024-00398830, RS-2020-II201361 and No.RS-2025-25441838) and the National Research Foundation of Korea (NRF) grant funded by the Korea government (MSIT) (RS-2024-00338439).
% Our method is designed to model semi-rigid deformations which are not major in motions of loose garments which are less influenced by body poses.
% Second, our current model does not encompass facial expressions. We leave these extensions as future work, where our framework can be enhanced by integrating priors from existing face~\cite{feng2021learning,danvevcek2022emoca} reconstruction methods. %and hand~\cite{pavlakos2024reconstructing} 
%following ExAvatar.

%We remain them as future works.

%We focus on the semi-rigid deformation, so we 

%% file: sec_camera_ready/supp.tex
% ---------------------------------------------------------------
% suppl.tex (문서 설정 없이 내용만 포함)
% ---------------------------------------------------------------

\section{Demonstration of Avatar Animation}
\begin{figure}[ht!]
  \centering 
  \includegraphics[width=\linewidth]{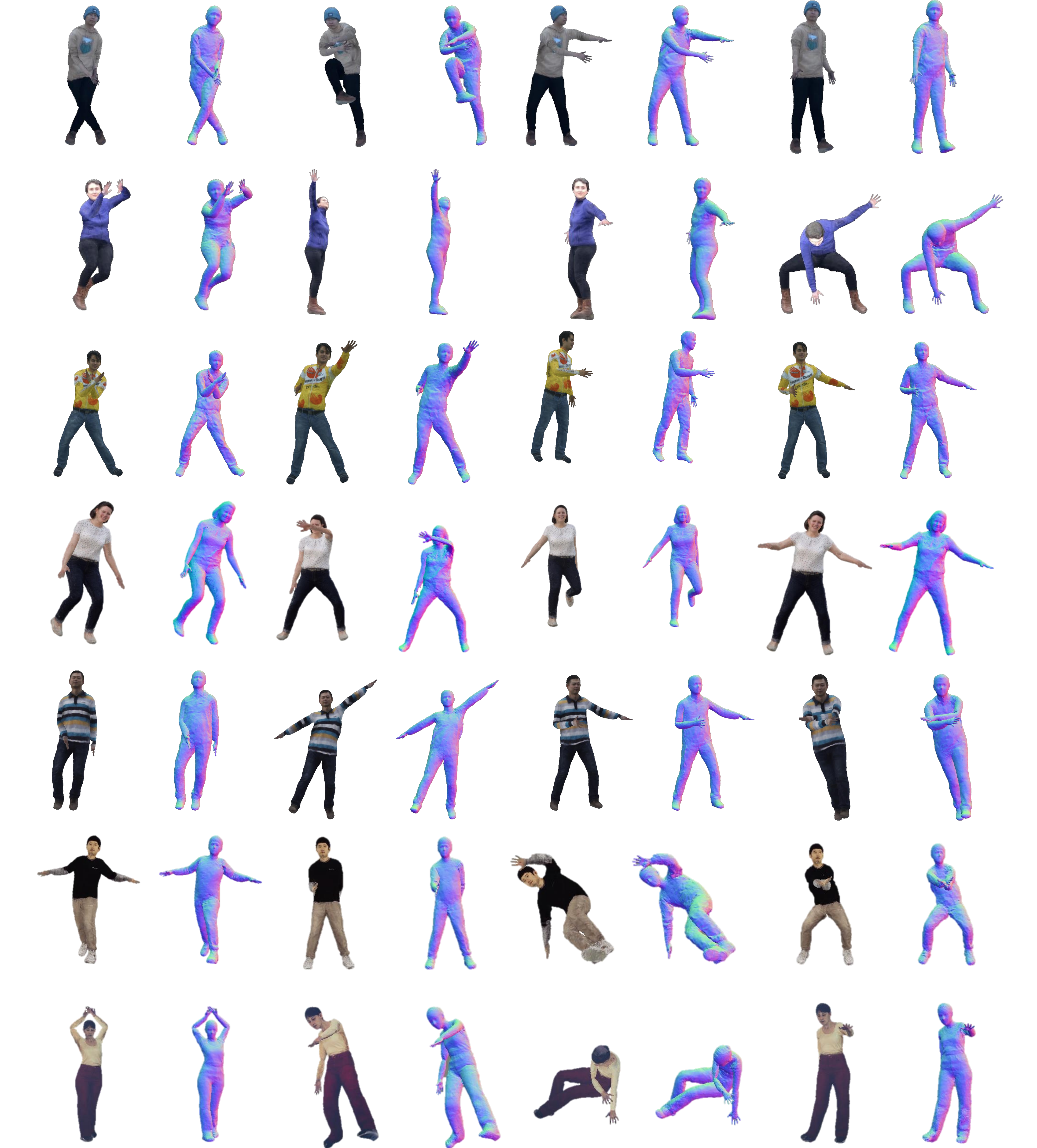}
  \caption{Animation results of our reconstructed avatars on the NeuMan, MonoPerfCap, and SynWild datasets.}
  \label{fig:animation_imgs}
\end{figure}

We provide a supplementary video showcasing the animation capabilities of our method. To generate realistic movements, we utilize SMPL body poses sampled from the AIST++ Dance Motion Dataset~\cite{li2021ai}. In addition to the supplementary video, we visualize the rendering results of the animated avatar, including both RGB images and normal maps in Fig.~\ref{fig:animation_imgs}. Furthermore, to demonstrate the generalization of our pipeline, we present animation results on the PeopleSnapshot dataset~\cite{alldieck2018detailed}, as shown in the supplementary video and Fig.~\ref{fig:peoplesnapshot_imgs}. In these figures, our reconstructed avatars exhibit plausible geometry and stable rendering quality.

\begin{figure}[ht!]
  \centering 
  \includegraphics[width=\linewidth]{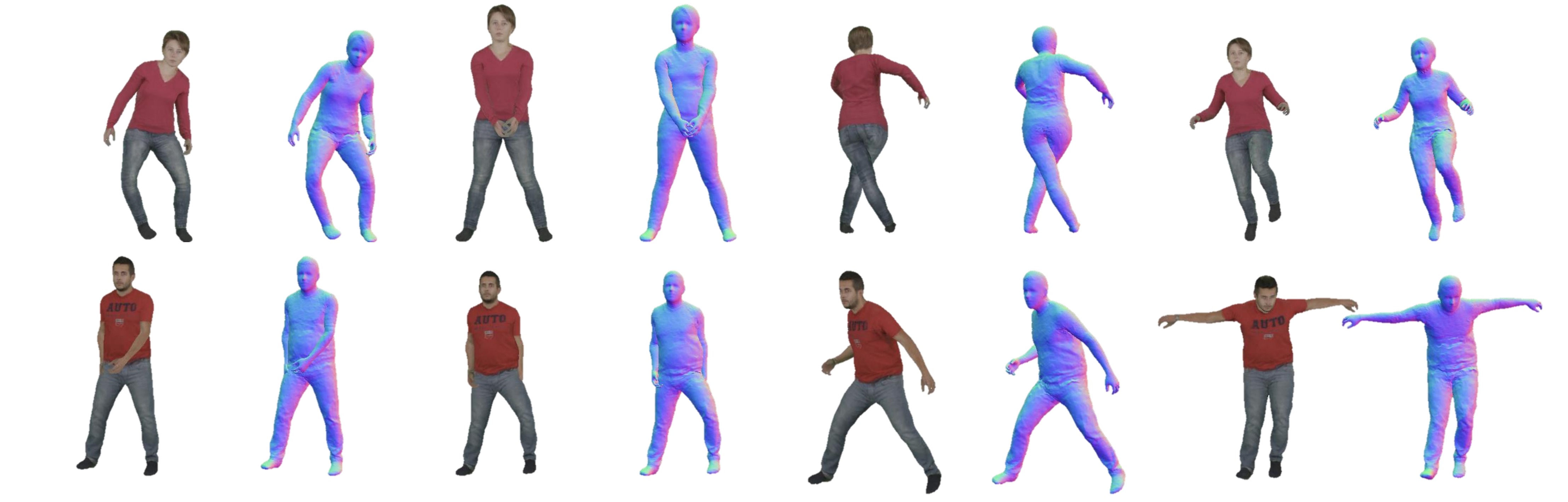}
  \caption{Animation results of our reconstructed avatars on the PeopleSnapshot dataset.}
  \label{fig:peoplesnapshot_imgs}
\end{figure}

\begin{figure}[ht]
  \centering 
  \includegraphics[width=\linewidth]{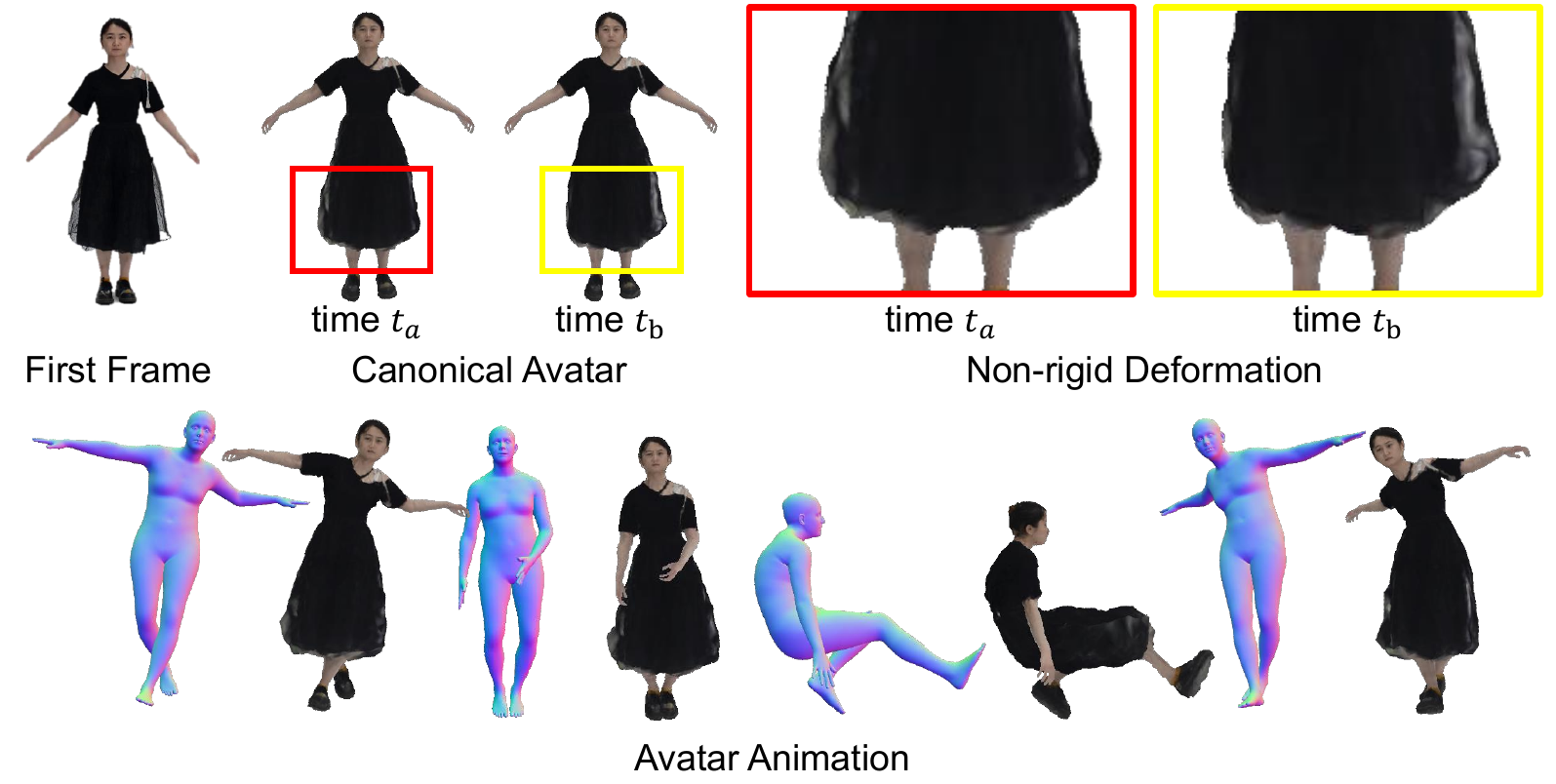}
  \caption{Reconstruction results of an avatar in loose clothing. Our pipeline achieves visually plausible results, even though it is primarily designed for semi-rigid deformations.}
  \label{fig:Loose_clothing}
\end{figure}

\section{Demonstration of Avatars in Loose Clothing}
As mentioned in the Conclusion section of the main paper, reconstructing avatars in loose clothing is out of the scope of our work. However, the coarse-to-fine manner of our pipeline can provide reasonable reconstruction quality even for such challenging cases. To demonstrate this capability, we reconstruct an avatar on a video sequence from SCARF~\cite{SCARF}, as shown in the Fig.~\ref{fig:Loose_clothing}. Here, our method successfully reconstructs the avatar and captures some non-rigid deformations, such as the dynamics of the dress. Furthermore, the reconstructed avatar shows stable animation results under unseen poses.

\section{Demonstration of pose driven semi-rigid deformation}
\begin{figure}[ht]
  \centering 
  \includegraphics[width=\linewidth]{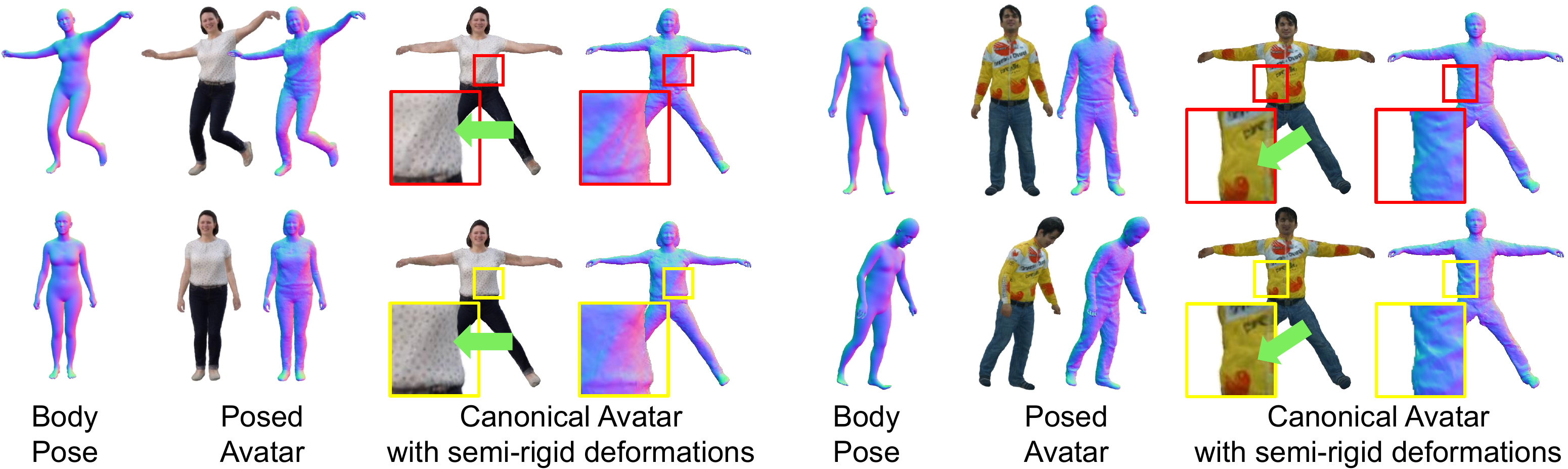}
  \caption{Visual demonstration of semi-rigid deformations driven by body poses of the avatars.}
  \label{fig:semi-rigid deformations}
\end{figure}
    
To capture realistic body dynamics, we formulate the semi-rigid deformation as a pose-dependent function. As shown in Fig.~\ref{fig:semi-rigid deformations}, we visualize these deformations on the canonical avatar to illustrate how they respond to varying body poses. For instance, when the right arm is raised (left panel), the resulting semi-rigid deformation naturally mimics the tension in the avatar's clothing.

\section{Additional Qualitative Results}
\begin{figure}[ht]
\centering
\includegraphics[width=\linewidth]{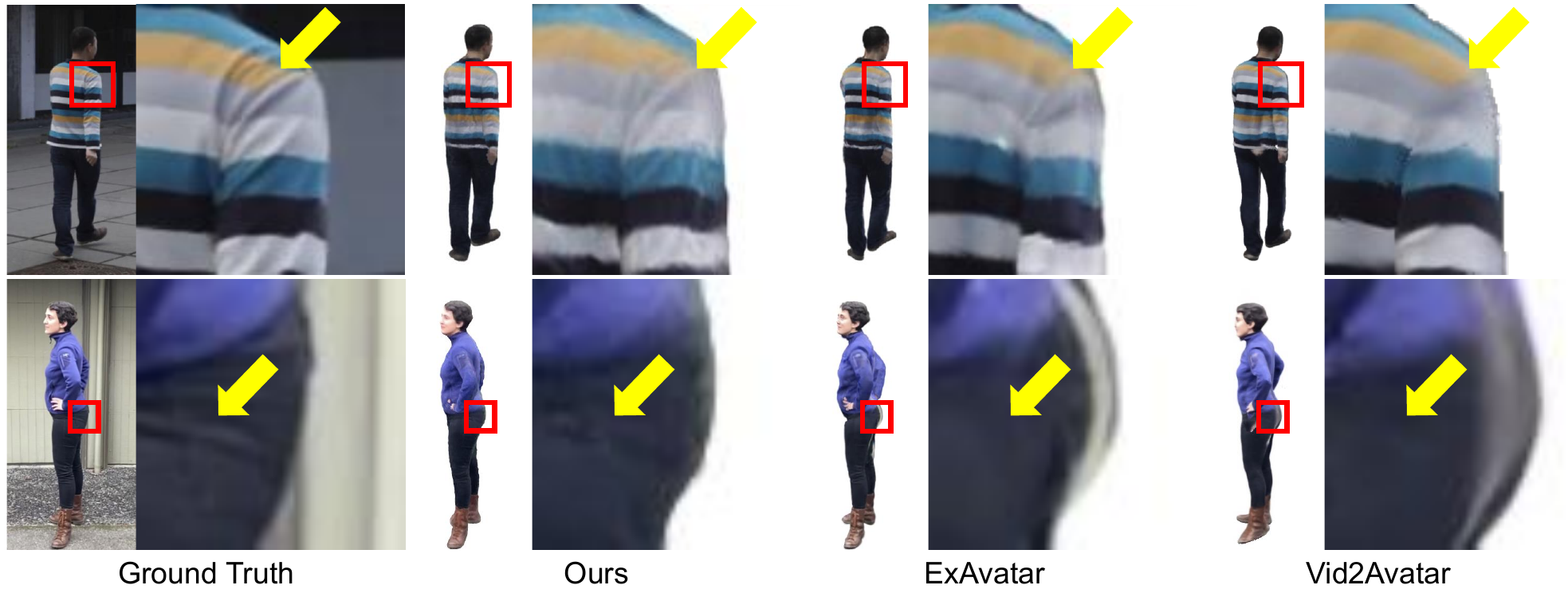}
\caption{Additional qualitative comparisons with ExAvatar and Vid2Avatar on the MonoPerfCap and NeuMan datasets.}
\label{fig:additional_qualitative}
\end{figure}

In Fig.~\ref{fig:additional_qualitative}, we provide additional qualitative comparisons on the test sets of NeuMan~\cite{jiang2022neuman} and MonoPerfCap~\cite{xu2018monoperfcap} datasets. Specifically, we compare our approach against ExAvatar and Vid2Avatar, which demonstrated state-of-the-art performance in the main paper. For a fair comparison on the NeuMan dataset, we utilize the results provided on their GitHub repositories. As shown in the figure, our pipeline reconstructs more realistic wrinkles compared to these baseline methods.

\section{Implementation Details}
\subsection{Details for Network of Jacobian fields}
In the original Neural Jacobian Fields (NJF), a single MLP predicts a 9-channel output that is directly reshaped into a $3 \times 3$ Jacobian matrix. However, this direct representation may introduce high non-linearity into the output matrix, making the optimization process more challenging. To alleviate this, we employ three separate 3-layer MLPs to independently predict the scale, shear, and rotation matrices, and then obtain the final Jacobian field by multiplying these components together. 

Specifically, the first MLP outputs a 6D continuous rotation representation~\cite{zhou2019continuity}, which is mapped to a $3 \times 3$ rotation matrix. The second and third MLPs estimate a 3D scale vector and a 3D shear vector, respectively. These scale and shear values are explicitly assigned to the diagonal and upper off-diagonal elements of an upper triangular matrix. Finally, we obtain the full Jacobian field by multiplying the rotation matrix with this upper triangular matrix.

\begin{table}[ht!]
\centering
\caption{Hyperparameters for our loss functions. We report the loss weights applied during the Coarse, Fine, and GS optimization stages.}
\label{tab:hyperparams}
\begin{tabular}{llccc}
\toprule
\textbf{Loss Term} & \textbf{Symbol} & \textbf{Coarse} & \textbf{Fine} & \textbf{GS} \\
\midrule
RGB L1 Loss                                     & $\lambda_{L1}$                        & $10.0$    & $10.0$    & $10.0$ \\
LPIPS Loss                                      & $\lambda_{lpips}$                     & $2.0$     & $2.0$     & $2.0$  \\
SSIM Loss                                       & $\lambda_{ssim}$                      & $2.0$     & $2.0$     & $2.0$  \\
Deformation-guided Residual Flow Loss           & $\lambda_{residual}$                  & $10.0$    & $20.0$    & $20.0$ \\
\midrule
L1 Regularization for Mesh Deformation          & $\lambda_{reg}$                       & $1.0$     & $1.0$     & -      \\
SDF-based Jacobian Regularization               & $\lambda_{SDF}$                       & $1.0$     & $0.1$     & -      \\
Dual Laplacian Regularization for $\mathbf{C}$  & $\lambda_{\mathbf{C}}$                & $1.0$     & $0.1$     & -      \\
Dual Laplacian Regularization for $\mathbf{C}^b$& $\lambda_{\mathbf{C}^b}$              & $1.0$     & $0.1$     & -      \\
Laplacian Regularization for $\mathcal{V}$      & $\lambda_{\text{lap}_\mathcal{V}}$   & $5{,}000$ & $30{,}000$& -      \\
\midrule
L2 Regularization for $\mathbf{s}$              & $\lambda_{\text{scale}}$              & -         & -         & $10.0$  \\
L2 Regularization for $\delta\boldsymbol{\mu}$               & $\lambda_{\text{offset}}$             & -         & -         & $1.0$ \\
Dual Laplacian Regularization for $\mathbf{s}$  & $\lambda_{\text{lap}\bm{s}}$   & -         & -         & $5{,}000$\\
Dual Laplacian Regularization for $\delta\boldsymbol{\mu}$ & $\lambda_{\text{lap}\delta \bm{\mu}}$    & -         & -         & $10.0$ \\
\bottomrule
\end{tabular}
\end{table}

\subsection{Hyperparameters}
In our framework, we employ various loss functions to optimize the avatar. The hyperparameters used across the different stages (Coarse, Fine, and GS) are detailed in Tab.~\ref{tab:hyperparams}.